\documentclass[journal]{IEEEtran}

\ifCLASSINFOpdf
   
\else

\fi
\usepackage[pdftex]{graphicx}
\usepackage{amsmath}
\usepackage{subfig}
\usepackage{xcolor}
\usepackage{comment}

\providecommand{\keywords}[1]
{
  \textbf{\textit{Keywords---}} #1
}
 
\begin{document}

\title{Not all domains are equally complex:\\Adaptive Multi-Domain Learning}



\author{\IEEEauthorblockN{Ali Senhaji\IEEEauthorrefmark{1},
Jenni Raitoharju\IEEEauthorrefmark{2}, Moncef Gabbouj\IEEEauthorrefmark{1} and
Alexandros Iosifidis\IEEEauthorrefmark{3}}\\
\IEEEauthorblockA{\IEEEauthorrefmark{1}Department of Computing Sciences, Tampere University, Finland\\
\IEEEauthorrefmark{2}Programme for Environmental Information, Finnish Environment Institute, Finland\\
\IEEEauthorrefmark{3}Department of Engineering, Aarhus University, Denmark\\
Emails: \IEEEauthorrefmark{1}ali.senhaji@tuni.fi,
\IEEEauthorrefmark{2}jenni.raitoharju@environment.fi,
\IEEEauthorrefmark{1}moncef.gabbouj@tuni.fi,
\IEEEauthorrefmark{3}ai@eng.au.dk}}

\maketitle

\begin{abstract}
Deep learning approaches are highly specialized and require training separate models for different tasks. Multi-domain learning looks at ways to learn a multitude of different tasks, each coming from a different domain, at once. The most common approach in multi-domain learning is to form a domain agnostic model, the parameters of which are shared among all domains, and learn a small number of extra domain-specific parameters for each individual new domain. However, different domains come with different levels of difficulty; parameterizing the models of all domains using an augmented version of the domain agnostic model leads to unnecessarily inefficient solutions, especially for easy to solve tasks. We propose an adaptive parameterization approach to deep neural networks for multi-domain learning. The proposed approach performs on par with the original approach while reducing by far the number of parameters, leading to efficient multi-domain learning solutions. 
\end{abstract}


\IEEEpeerreviewmaketitle

\section{Introduction}
Deep learning is achieving state-of-the-art results in a wide range of applications. Particularly in computer vision, deep learning has reached astonishing performance compared to traditional methods in nearly all problems \cite{krizhevsky2012imagenet,lecun2015deep,gu2018recent}. However, most of these solutions aim to solve one specific task. For $N$ different tasks on different domains, this archetype requires having $N$ models, one for each of the $N$ target tasks. The community is now challenging this established paradigm of training problem-specific models. The next goal is to learn complex independent problems jointly. Methods aiming at addressing this challenge come in two different flavors. The first one is \textit{multi-task learning} \cite{caruana1997multitask,liu2019end,doersch2017multi} aiming to develop an architecture able to extract diverse information from a given sample, e.g., an image. To illustrate this, we can think of vision systems in autonomous cars. These need to detect objects, perform semantic image segmentation to extract the location of road, detect pedestrians and other cars, as well as recognize road signs. The second flavor is \textit{multi-domain learning} \cite{bulat2019incremental,rebuffi2018efficient,rebuffi2017learning} focusing on achieving good results on a multitude of different visual domains. This means developing one architecture able to extract information from diverse domains, such as images from autonomous vehicles, space images, hand written text, medical images, or flower images.


In this work, we are concerned with multi-domain learning. In the literature, multi-domain learning has been framed in two different ways. The first one is based on the idea to learn an early-stage generic feature extractor, then branch to the different domains with more domain specific parameters \cite{wang2019towards}. This generic feature extractor should learn basic features that can be used as building blocks for solving more complex visual tasks (Fig. 1a). It can be trained from scratch or it can use transfer learning. However, such an approach does not perform as well as learning domain-specific architectures. The second approach uses a set of parameters to form a domain agnostic model. The parameters of that model are shared among all the domains while adding domain-specific parameters through its architecture to form models specialized to each domain (Fig. 1b) \cite{rebuffi2018efficient,rebuffi2017learning}. Because the domain agnostic model forms the basis for all the domain-specific models, it is referred to as the \textit{base model} hereafter. 

We can think of a deep neural network as a parametric function $\Phi(x; \mathcal{W}, \mathcal{A}_n)$, the parameters of which are trained for solving a (visual) task. Here $x$ denotes an input sample (e.g., an image), $\mathcal{W}$ contains the parameters of the base model and $\mathcal{A}_n$ contains domain-specific parameters for domain $n$. For a multi-domain model to perform well on $N$ domains, we would need $\sum_n |\mathcal{A}_n|$ domain specific parameters, where $|\cdot|$ denotes the cardinality of a set. The goal is to achieve high performance on all domains while adding a small number of additional parameters $\mathcal{A}_n, \:n=1,\dots,N$. Here, we should note that since the domain-specific parameters $\mathcal{A}_n$ are learned independently for each new domain $n$, one can incrementally train the domain-specific parameters for each individual domain without the need to store and use the training data for the remaining domains. This is why for solutions following the above-described approach the terms \textit{multi-domain learning} and \textit{incremental multi-domain learning} are used interchangeably in the literature. 

\begin{figure*}[htp]
    \centering
    \includegraphics[width=0.9\linewidth]{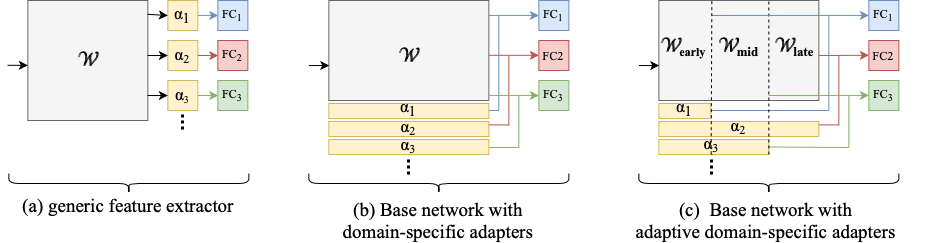}
    \caption{\textbf{Adaptive parametric base network.} We propose an adaptive parametric neural network architecture (c) instead of generic feature extractor (a) and adapting the whole base network (b).}
    \label{fig:galaxy}
\end{figure*}


A common approach followed by multi-domain learning methods is to augment the set of shared parameters with a small number of domain-specific parameters, i.e., $|\mathcal{A}_n| = q |\mathcal{W}|, \:n=1,\dots,N$, where $q \simeq 0.1$. This means that for solving each of the new tasks the topology of the domain-specific models is fixed and requires traversing the entire topology of the base model, while augmenting it using a small number of domain-specific parameters to specialize the response of the corresponding network in its intermediate layers to the specific domain. We argue that since different tasks coming from different domains have different complexities, the above-described approach leads to unnecessarily complex solutions, especially for easier tasks. For a human, it would be an easy task to detect a pedestrian in an image, read numbers, or classify characters. On the other hand, it is challenging to take a logical visual reasoning test or classify objects in fine-grained categories, such as flowers or aircrafts. Humans have a general representation of the visual world; some tasks require more detailed observation by looking at more complex features, while other tasks are easier and require less detailed observation. 

In this paper, we investigate the adaptive use of the base model parameters and we propose a novel adaptive parameterization approach for multi-domain architectures depending on the level of complexity of each individual domain (Fig. 1c). Depending on the difficulty of a new domain, the domain-specific model can partially exploit the base network topology to achieve results comparable to using the entire network topology, leading to high parameter reduction without sacrificing in performance. The use of compact and efficient models is important for a multitude of applications, like when using edge devices requiring compressed networks due to the limited computational power and memory. Our approach makes it more efficient to compress, swipe, and transfer the models.

\section{Related Work}
\textbf{Multi-task learning (MTL)} aims to concurrently learn a number of different tasks (e.g., segmentation, classification, detection) by sharing information to achieve better results. The first works in MTL \cite{caruana1997multitask} based on neural networks approached the problem sharing the first layers followed by specialized layers for the different tasks. It was shown in \cite{caruana1997multitask} that sharing the parameters across different tasks improves performance. Learning multiple tasks at the same time constraints the network to learn a generalized representation of the input data \cite{he2017mask}. MTL is concerned with designing one architecture to learn different tasks from the same domain, while requiring storing and using the data of all tasks concurrently to train the model. In some cases such an approach is infeasible, e.g., when the data from different tasks are becoming available over time and/or when the total size of data for all the tasks is enormous. 

\textbf{Domain adaptation (DA)} aims at tuning the parameters of a model trained on data following one distribution (\textit{source domain}) to \textit{adapt} so that they can provide high performance on new data following another distribution (\textit{target domain}). In \cite{ganin2014unsupervised,tzeng2015simultaneous}, deep neural network parameter adaptation is achieved by minimizing the discrepancy between data coming from the two domains. In \cite{li2018learning}, a meta-learning based method is introduced to make the model robust toward domain changes. The supervised DA problem is formulated as a graph embedding problem in \cite{hedegaard2020dage} which is subsequently optimized for deep neural networks. Here, we should note that a common application scenario, where supervised DA methods are used, requires the two domains to be formed by the same sets of classes. Moreover, after adapting the parameters of the model, performance of the resulting model is expected to drop for data coming from the source domain.

\textbf{(Incremental) Multi-domain learning (MDL)} deals with different sets of target classes from one domain to another. Moreover, once trained, the performance of the resulting model on each domain remains unchanged. Recent works \cite{rebuffi2018efficient,rebuffi2017learning,thewlis2017unsupervised} are proposing architecture designs to classify data from multiple domains incrementally. The main goal is to learn a single model that performs well across all target domains while minimizing the amount of additional domain-specific parameters. 
In \cite{bulat2019incremental}, an incremental multi-domain learning was introduced by jointly parametrizing weights throughout the layers using a low-rank Tucker structure. In \cite{sarwar2019incremental}, an incremental method was proposed: it is based on sharing some of the early layers and splitting later ones. Authors of \cite{rosenfeld2018incremental} introduced an incremental learning method by constraining newly learned filter banks to be a linear combinations of existing ones.
In \cite{thewlis2017unsupervised}, a single architecture is proposed that shares all parameters across the domains and learns domain specific batch normalization parameters. A network architecture extending the one in \cite{thewlis2017unsupervised} with additional parameters (adapters) which increase the number of the network parameters for each domain by around $10\%$ is proposed in \cite{rebuffi2017learning}. This method requires the domain specific parameters to be embedded with the network when the base model is trained. This limitation is tackled in \cite{rebuffi2018efficient} by introducing an efficient parameterization through a parallel configuration of the adapters. This approach makes it possible to plug the learned domain specific parameters to any pre-trained convolutional neural network (CNN) base model. We use this feature to parametrize specific blocks of the base model in an adaptive manner. 

\textbf{Early exits} are used as a way to deploy fast inference for deep networks. These networks come with a considerable computational cost, which limits their usability in many real-life application. 
In \cite{huang2017multi}, a method introducing early exits able to make a prediction at any point of the network is introduced. This method is useful in cases of budgeted batch classification.
In \cite{veit2018convolutional}, the authors propose a method to decide on the needed layers for each input image by estimating layer relevance to the sample complexity. BranchyNet \cite{teerapittayanon2016branchynet} proposes a method based on branch classifiers early in the network to classify easier samples.
An adaptive inference approach using neural bag-of-features to tackle some of the limitations of prior methods using early exits is proposed in \cite{passalis2019adaptive}. All the methods above still require all model parameters to make inference for the entire set of samples. Our adaptive parametric method brings a novel approach enabling the use of only a portion of the base model parameters to solve tasks in less challenging domains.



\section{Adaptive Multi-domain Learning}
\label{method}

\begin{figure}[!t]
    \centering
    \includegraphics[width=0.9\linewidth]{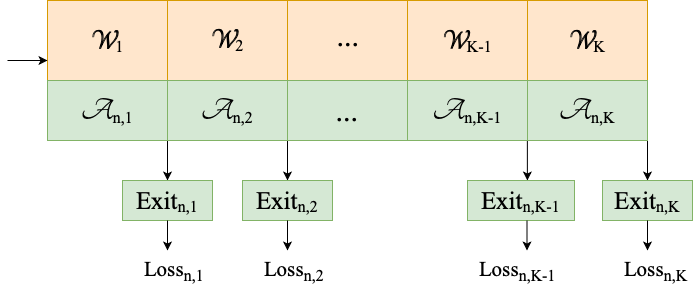}
    \caption{The partition of the base model into K non-overlapping blocks of layers adapted with $\alpha_n$. Exit modules are placed after every portion of the base model.}
    \label{fig:k-exits}
    \label{k-exits}
\end{figure}

We propose an adaptive parametric neural network architecture for multi-domain learning, where the complexity of the architecture for a specific domain depends on the level of complexity of each domain $\mathcal{D}_n, \:n=1,\dots,N$. We show that we do not need to use all the parameters of the base network $\mathcal{W}$ for achieving high performance in all the target domains. Within a convolutional neural network, basic features are learned in earlier layers and more complex features are learned in later layers of the network \cite{zeiler2014visualizing}. Thus, we can divide any deep convolutional neural network into $K$ non-overlapping blocks of layers, as shown in Fig. \ref{k-exits}. These blocks of the network are formed by a different set of parameters $\mathcal{W}_k \in \mathcal{W}$ corresponding to different sets of learned features having an increasing complexity as new blocks of the model are added. To obtain domain-specific outputs on each layer of the network, we inject a set of adapters $\mathcal{A}_{n,k}$ for every block $k$, where the first and the second subscripts denote the domain and the network block, respectively. The use of these adapted network blocks for modeling new domains allows for a better flexibility using the base model parameters. Easier domains will require the use of a smaller number of blocks of the network compared to more complex domains. The base network parameters $\mathcal{W} = \cup^{K}_{k=1} \mathcal{W}_k$ are frozen for all the domains, we only learn domain specific parameters $\mathcal{A}_n$. When a domain-specific model exploits and adapts $K_n$ blocks of the base model, the overall number of parameters for that model is equal to $|\mathcal{W}_n| + |\mathcal{A}_{n}|$, where $\mathcal{W}_n = \cup^{K_n}_{k=1} \mathcal{W}_k$ and $\mathcal{A}_{n} = \cup^{K_n}_{k=1} \mathcal{A}_{n,k}$.

\begin{figure}[tpb]
    \centering
    \includegraphics[width=0.9\linewidth]{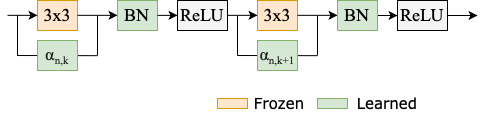}
    \caption{The parallel configuration of domain specific adapters within two consecutive convolution layers. }
    \label{fig:parallel}
    \label{parallel}
\end{figure}

The basic building block of a deep CNN is the convolution layer $\mathcal{C}$ followed by an activation function. The network can be thought of as a chain of convolution layers $\mathcal{C}_1 \circ \dots  \circ \mathcal{C}_n$. Let us denote by $W_k$ the convolution filters of layer $k$. A layer can be expressed as $\mathcal{C}_k(x;W_k)=  g(W_k * x)$, where $x$ is the (tensor) input to the layer, $*$ denotes convolution, and $g(\cdot)$ is the activation function (e.g. Rectified Linear Unit) applied in an element-wise manner. Injecting domain specific learnable filters, as displayed in Fig. \ref{fig:parallel}, can be represented by adding parallel convolutions with $\alpha_{n,k}$ parameters:
\begin{equation}
\mathcal{C}_{k,adap}(x;W_k,\alpha_{n,k}) = g\Big( (W_k*x)  +  (\alpha_{n,k}*x) \Big).
\label{eq_adp}
\end{equation}
This formulation requires the outputs of the convolution with the base model parameters $(W_k*.)$ and the convolution with the newly introduced adapters $\alpha_{n,k}$ to have the same output size for each layer $l$ to make the addition possible. This is obtained by setting a number of adapters equal to the number of convolution filters for each convolution layer. Here, we should note that even though in Eq.(\ref{eq_adp}) the blocks parameterized using the adapters $\alpha_{n,k}$ correspond to individual layers, it is straightforward to parameterize multiple consecutive layers.

\begin{figure*}[htpb]
    \centering
    \includegraphics[width=0.95\linewidth]{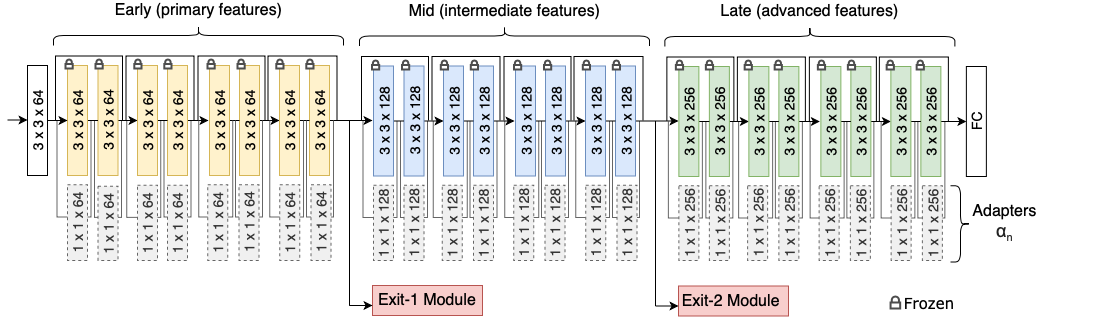}
    \caption{A ResNet architecture with a 26-layer configuration. Domain agnostic filters are pre-trained and frozen. Domain specific $1 \times 1$ filter banks are added in parallel to the residual units. Two exit modules are placed after the early and mid blocks.}
    \label{fig:archi}
\end{figure*}

After training the base model, when a domain-specific network is trained, the $c \times c$ convolution filters of the base model layer used for all domains are frozen. For each block of a new target domain, we learn an adapter of size $1 \times 1$, as shown in Fig \ref{parallel}\footnote{The filters of both the base model and the domain-specific adapters are tensors with the same number of channels, i.e., $c \times c \times I$ and $1 \times 1 \times I$, respectively, where $I$ is the number of channels of the input tensor to that layer.}. Batch normalization parameters $\mu_{n,k}$ and $\sigma_{n,k}, \:k=1,\dots,K$ are learned for each domain $n$ as well. For each domain $n$, we introduce early exit modules (using dense layers or convolution layers), each receiving as input the output of a block $k$. We learn the domain-specific parameters for the $k$-th block  $\mathcal{A}_{n,k} = \{\alpha_{n,k},\mu_{n,k},\sigma_{n,k}\}$ and the additional exit modules parameters through Backpropagation by minimizing the cross-entropy loss:
\begin{equation}
\mathcal{L}_k = -\sum_{x_j \in \mathcal{D}_n} \Bigg( \sum_{e=k}^{K} \sum_{i=1}^{C_n}y_{j,i}*log(f_{e,i}(x_{j})) \Bigg),
\label{losses}
\end{equation}
where $x_j$ is a sample from domain $\mathcal{D}_n$, $C_n$ is the number of classes in $\mathcal{D}_n$, $y_j$ is the one-hot encoded ground truth label vector corresponding tos ample  $x_j$, and $f_e(.)$ is the softmax output of domain-specific model at exit $e$. That is, the domain-specific parameters of the $k$-th block are updated based on the errors obtained from all layers following block $k$.

To decide which exit of the domain-specific model to use  for domain $\mathcal{D}_n$ (and thus define the final number of blocks $K_n$ to be used for that domain) we evaluate the performance at each exit of the adapted model. A threshold $T_n$ is used to express the acceptable drop in performance when choosing an early exit, compared to the performance obtained when using the entire (adapted) model.


To illustrate our approach, let us consider a ResNet-26 architecture \cite{he2016identity} as a the base network formed by the parameters $\mathcal{W}$ to be shared across the different domains. The network can be parametrized using residual adapters \cite{rebuffi2018efficient,rebuffi2017learning}. Residual adapters show capacity to adapt the base model parameter set $\mathcal{W}$ for a multitude of visually distinct domains. These adapters are domain-specific parameters added to the network in the form of $1 \times 1$ filter banks. For two consecutive convolution layers with a kernel size of $c \times c$ and a fixed number of input and output channels $I$, the number of base model parameters shared among all domains is $2(c^2I^2 + I$), where the multiplication by $2$ accounts for the two $c \times c$ convolution filter banks and the additional $I$ accounts for biases. For the domain-specific $1 \times 1$ adapters, the number of parameters per residual unit is $2(I^2 + 3I$), where $3I$ accounts for the biases and batch normalization parameters. When a value of $c=3$ is used, the domain-specific parameters are about $9$ times less than the number of the base model parameters.

We can divide the network into three different blocks: an `early' block which is able to encode primary features, a `mid' block which can encode intermediate features, and the `late' block which can encode advanced and more complex features as shown in Fig. \ref{fig:archi}. This choice leads to adding two exit modules to the resulting architecture, one after the `early' block and another one after the `mid' block. The adaptive parameterization described above can lead to the following three choices for a new domain: the adoption of an adapted `early' block of the base model can lead to high performance when the corresponding problem is easy. This requires using only $\sim$4.7\% of the base model parameters and parameters it would have taken to adapt the entire base network. For a domain with  intermediate level of complexity, $\sim$23.8\% of the base model parameters and parameters it would have taken to adapt the entire base network are needed. For challenging domains, the whole base network and domain specific adapters are used. In each case, the parameters needed for adapting the network account for $\sim$10\% of the corresponding base model parameters.



\section{Experiments}


We used residual network (ResNet) \cite{he2016identity} as a base model for our experiments. We chose a 26-layer configuration, making a fair compromise in speed and learning capacity of the network. The model has three macro-blocks, each having a four residual units with two convolution layers (3 x 3 filters) and a skip connection. Before the first macro-block, there is one convolutional layer. After the last macro-block there is a fully-connected classification layer. For every domain this last layer is replaced to fit the number of classes of the specific domain. The network takes $72 \times 72 \times 3$ (rescaled) images. Macro-blocks output 64, 128 and 256 feature channels in this order. The resolution drops by half from a macro-block to another. We used the macro-block division to split the network into an `early', `mid', and `late' blocks. These splits embed primary, intermediate, and advanced features, respectively. The domain agnostic parameters were trained on ImageNet at the beginning and then frozen. We parametrized the ResNet-26 architecture using the $1 \times 1$ filter banks discussed in Section \ref{method}. Fig. \ref{fig:archi} shows the adapters configured in a parallel fashion with each residual unit. Two exit modules were placed after the early and mid macro-blocks. For the last (third) exit, we used the basic exit module with only a fully connected layer in all our experiments. The architecture shown in Fig. \ref{fig:archi} is fully used during training. We report results from the three exits for all the domains. 

\begin{figure}[tp]
    \centering
    \includegraphics[width=1\linewidth]{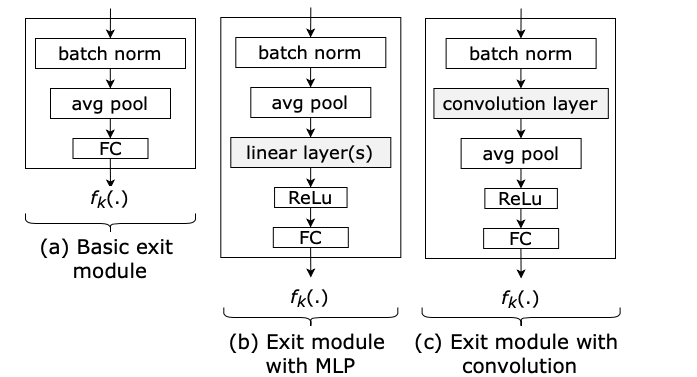}
    \caption{The different exit topologies used in the early and mid exits: (a) Basic exit module, is similar to the tail layer of the original architecture. (b) Exit module with MLP has an additional layer or two followed by an activation function (ReLu). (c) Exit module with convolution has an additional convolution layer followed by an activation function (ReLu).}
    \label{fig:galaxy}
    \label{exits}
\end{figure}

We investigated the three different exit module topologies shown in Fig.~\ref{exits}. The basic exit module Fig.~\ref{exits}(a) has no extra learnable parameters beside domain specific $\mu_n$ and $\sigma_n$ for the batch normalization. These batch normalization parameters were learned for the two other modules -Fig.~\ref{exits}(b) and  Fig.~\ref{exits}(c)- as well. Using the basic module, we wanted to assess the capacity of the adapters alone to parametrize the different blocks of the network. The basic module is similar to the classification layer for the third exit. Then we had an exit module with a multi-layer perceptron (MLP) shown in Fig.~\ref{exits}(b). The goal is to learn extra layers before the final classification layer. We investigate single layers with 128 and 512 neurons, as well as two fully connected layers with 128 neurons each. The MLP exit module is meant to play the role of a classifier based on the given features from the different blocks of the network. For the last exit module topology shown in Fig.~\ref{exits}(c), we use a convolution layer with minimal $1 \times 1$ convolution filter banks. The rational behind investigating this topology was to enable learning missing features for each domain.

In our experiments, we used the Visual Decathlon Benchmark proposed in \cite{rebuffi2017learning}. The benchmark combines ten different well-known datasets from different visual domains, statistics of which are illustrated in Table \ref{dataset}. These datasets represent a wide range of domains from classifying airplanes, detecting pedestrians, classifying flowers to action recognition from realistic action videos. The number of target classes by domain varies from a domain to another, as well as the number of images per dataset as shown in Table \ref{dataset}.

\begin{table}
\setlength{\tabcolsep}{2.7pt}
\caption{The train/val/test split sizes for the ten different datasets from multiple visual domains, as well as the number of classes per domain. }
\label{dataset}
\centering
\renewcommand{\arraystretch}{1.2}
\begin{tabular}{l|ccccc}
Dataset     & no. classes & training & validation & testing \\
\hline
Airc         & 100         & 3334     & 3333       & 3333    \\
C100         & 100         & 40000    & 10000      & 10000   \\
DPed        & 2           & 23520    & 5880       & 19600   \\
DTD       & 47          & 1880     & 1880       & 1880    \\
GTSRB             & 43          & 31367    & 7842       & 12630   \\
ImNet          & 1000        & 1232167  & 49000      & 48238   \\
OGlt          & 1623        & 19476    & 6492       & 6492    \\
SVHN              & 10          & 47217    & 26040      & 26032   \\
UCF         & 101         & 7629     & 1908       & 3783    \\
Flwr       & 102         & 1020     & 1020       & 6149   
\end{tabular}
\\[10pt]
\end{table}

\begin{table*}[]
\small
\setlength{\tabcolsep}{6.0pt}
\centering
\caption{Classification accuracy (top-1) for the three exits (early-1, mid-2, late-3) for each experiment.}
\renewcommand{\arraystretch}{1.0}
\begin{tabular}{l|c|c|cccccccccc|c} 
Method                   & \%P. & \#P. & ImNet & Airc. & C100  & DPed  & DTD   & GTSR  & Flwr  & OGlt  & SVHN  & UCF   & mean  \\  \hline
Finetune \cite{rebuffi2018efficient}                & 10x & 61.94M  & 59.87 & 60.34 & 82.12 & 92.82 & 55.53 & 97.53 & 81.41 & 87.69 & 96.55 & 51.20  & 76.51 \\   \hline
A. Series \cite{rebuffi2017learning}        & 2x  & 12.38M  & 59.67 & 56.68 & 81.20  & 93.88 & 50.85 & 97.05 & 66.24 & 89.62 & 96.13 & 47.45 & 73.88 \\ 
A. Paral \cite{rebuffi2018efficient}   & 2x   & 12.38M & 60.32 & 50.29 & 81.01 & 90.57 & 51.65 & 99.02 & 70.24 & 87.42 & 95.84 & 48.01 & 73.44 \\ \hline

No-adp E-1           & 0.05x & 296.6K & -     & 10.12 & 25.97 & 66.58 & 24.74 & 41.18 & 33.03 & 41.34 & 27.39 & 14.99 & 31.71 \\
No-adp E-2            & 0.24x &  1.47M & -     & 7.30   & 20.08 & 64.64 & 17.88 & 29.55 & 21.18 & 49.12 & 25.05 & 11.48 & 27.37 \\
No-adp E-3            & 1x   & 6.19M & 60.32 & 4.33  & 9.25  & 63.52 & 11.71 & 19.02 & 10.61 & 21.67 & 20.84 & 6.96  & 22.83 \\ \hline

Basic E-1            & 0.09x  & 591.5K & -     & 16.30  & 53.85 & 87.05 & 39.26 & 96.28 & 47.87 & 81.79 & 89.72 & 31.09 & 60.36 \\
Basic E-2            & 0.47x & 2.95M & -     & 42.55 & 72.20  & 89.86 & 48.20  & 97.88 & 56.31 & \textcolor{blue}{87.13} & \textcolor{blue}{95.35} & 43.30  & 70.31 \\
Basic E-3         & 2x   & 12.38M & 60.32 & 46.18 & 78.00    & 89.84 & 49.53 & 97.86 & 59.19 & 87.41 & 96.06 & 46.95 & 71.14 \\ \hline

MLP128 E-1           & 0.09x & 599.7K & -     & 24.07 & 53.65 & \textcolor{blue}{87.33} & 43.73 & \textcolor{blue}{97.00 }   & 54.18 & 80.95 & \textcolor{blue}{90.04} & 30.19 & \textcolor{blue}{62.35} \\
MLP128 E-2            & 0.48x  & 2.95M & -     & 45.04 & 71.85 & 90.80  & \textcolor{blue}{49.15} & 98.52 & 63.14 & 86.44 & \textcolor{blue}{95.24} & \textcolor{blue}{43.12} & \textcolor{blue}{71.48} \\
MLP128 E-3                 & 2x  & 12.38M  & 60.32 & 49.36 & 78.55 & 90.93 & 49.90  & 98.86 & 68.06 & 88.32 & 96.21 & \textcolor{blue}{48.64} & 72.92 \\ \hline

MLP128-B E-1 & 0.09x & 599.7K  & -     & 23.23 & 56.90  & \textcolor{blue}{87.72} & 39.85 & 95.77 & 46.32 & 80.71 & 90.16 & 33.15 & 61.54 \\
MLP128-B E-2 & 0.48x  & 2.95M & -     & 43.63 & 71.88 & 89.74 & 47.56 & 96.87 & 58.42 & 85.70  & 95.02 & 43.33 & 70.24 \\
MLP128-B E-3        & 2x  & 12.38M  & 60.32 & \textcolor{blue}{50.29} & \textcolor{blue}{81.01} & 90.57 & 51.65 & 99.02 & \textcolor{blue}{70.24} & 87.42 & 95.84 & 48.01 & 73.44 \\ \hline

MLP512 E-1            & 0.10x & 624.3K & -     & 24.40  & 54.60  & 87.08 & 40.64 & 95.81 & 53.67 & 80.97 & 87.25 & 31.96 & 61.82 \\
MLP512 E-2            & 0.48x & 3.01M & -     & 45.37 & 71.67 & 90.78 & 47.40  & 97.78 & 61.87 & 86.46 & 95.30  & 44.20  & 71.21 \\
MLP512 E-3            & 2x  &  12.38M & 60.32 & 50.32 & 78.18 & 90.93 & 48.73 & 97.78 & 64.73 & 87.78 & 96.28 & 46.90  & 72.20  \\ \hline

MLP-2L E-1            & 0.10x & 616.1K & -     & 23.17 & 53.04 & 86.86 & 41.07 & 95.54 & 53.93 & 74.62 & 87.67 & 30.06 & 60.67 \\
MLP-2L E-2            & 0.48x & 2.98M & -     & 44.74 & 70.49 & 90.46 & 47.98 & 96.88 & 63.97 & 83.80  & 94.78 & 42.64 & 70.64 \\
MLP-2L E-3            & 2x   & 12.38M & 60.32 & \textcolor{blue}{50.62} & 77.94 & 90.68 & 48.78 & 97.46 & 69.85 & 88.34 & 96.06 & \textcolor{blue}{49.04} & 72.91 \\ \hline

CL E-1                  & 0.09x & 595.6K & -     & 23.47 & 52.55 & 86.56 & 41.18 & 96.16 & 51.66 & 80.79 & 88.60  & 30.91 & 61.32 \\
CL E-2                  & 0.47x & 2.96M & -     & 43.21 & 71.34 & 89.89 & \textcolor{blue}{49.53} & 97.76 & 59.53 & 86.20  & 95.04 & 43.33 & 70.65 \\
CL E-3                  & 2x &  12.38M  & 60.32 & 47.74 & 77.50  & 90.14 & 49.10  & 97.81 & 63.23 & 87.94 & 96.16 & 47.11 & 71.71 \\ \hline \hline
Best T=3.5\%                  & 1.53x &  9.50M  & 60.32 & 50.62 & 81.01  & 87.72 & 49.53  & 97.00 & 70.24 & 87.13 & 95.35 & 49.04 & 72.79 \\ 
\end{tabular}
\label{tab:results}
\end{table*} 

\begin{figure}[!t]
\centering
\subfloat[Example of the easy domain: GTSRB dataset.]{\includegraphics[width=1\linewidth]{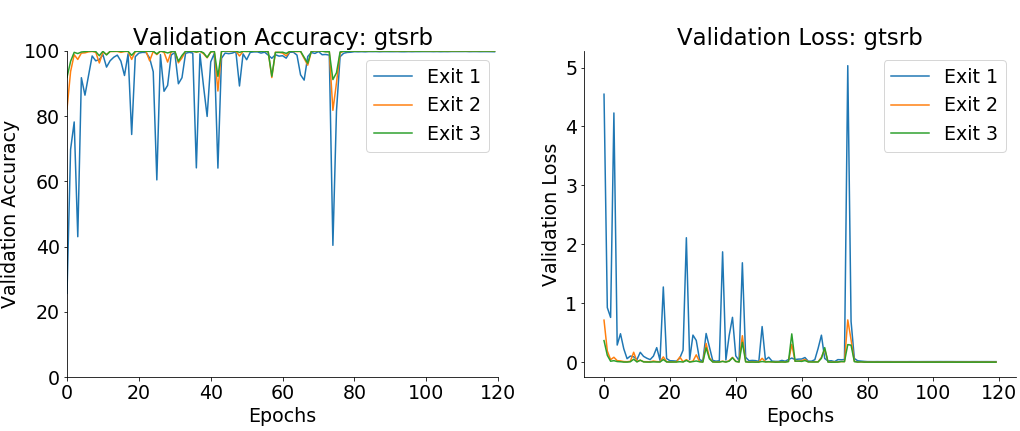}
\label{gtsrb}}
 \hfil \\
\subfloat[Example of the intermediate domain: SVHN dataset.]{\includegraphics[width=1\linewidth]{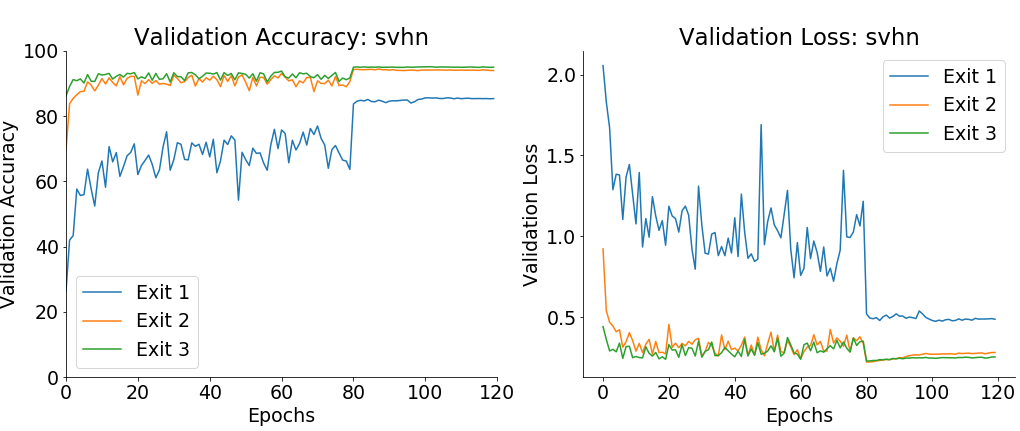}
\label{svhn}}
 \hfil \\
\subfloat[Example of the challenging domain: VGG-Flowers dataset.]{\includegraphics[width=1\linewidth]{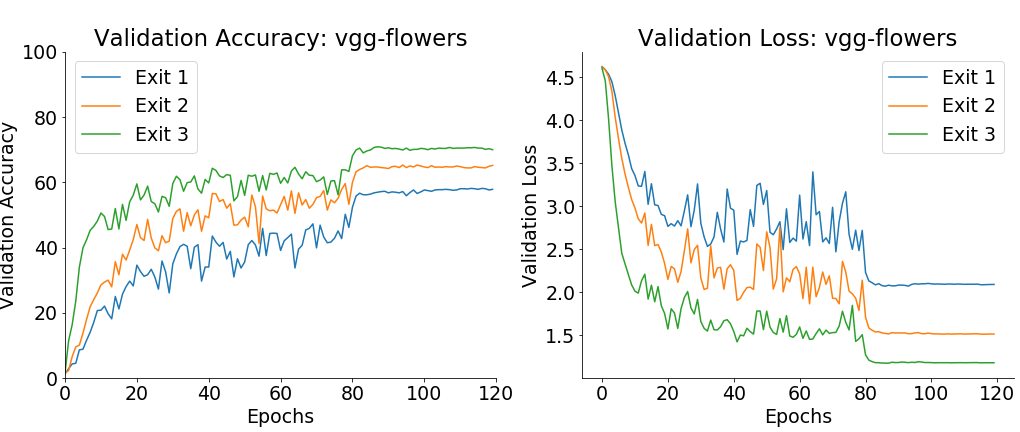}
\label{vgg}}
\caption{Accuracy and loss plots on the validation set during training, using an exit module with a 128 linear layer, over 120 epochs, forthree domains of different complexity levels.}
\label{fig_sim}
\end{figure}


First we started by taking a pretrained base network -domain agnostic parameters- on ImageNet, then it was frozen. As a baseline, we trained the residual adapters in a parallel configuration from scratch as described in \cite{rebuffi2018efficient}. Despite using the codes provided by the authors of \cite{rebuffi2018efficient}, our results are somewhat different from those published in \cite{rebuffi2018efficient}. In our comparisons, we use the results obtained by ourselves to ensure a fair comparison. 

We trained an independent set of adapters $\cup^{3}_{k=1} \alpha_{n,k}$ for every domain. Domain specific parameters were learned using stochastic gradient descent. We trained for 120 epochs, starting with a learning rate of 0.1, we decreased it to 0.01 starting at epoch 80, then to 0.001 at epoch 100. We used different weight decays depending on dataset size. We set a higher weigh decay for smaller datasets and smaller one for bigger dataset as described in \cite{rebuffi2018efficient}.  This adaptive weight decay setting plays the role of regularization for smaller datasets to prevent over-fitting. The same weight decays were used throughout the experiments. The whole base network was used to infer images from the ImageNet domain with no extra adaptation. 

To understand the importance of different factors of the proposed approach, we performed the following experiments: `MLP128' and `MLP512' are experiments with early exit modules having one dense layer of size 128 and 512, respectively. `MLP-2L' is an experiment where the early exit modules have two dense layers of size 128. In `MLP128-B', we had the same setup as as in `MLP128' but we followed a different training strategy from the one described in Section \ref{method}. We trained the different parts block-wise. In the first exit, we  trained the adapters for the early part of the network stand alone based only on the loss from the first exit. In the second exit, we trained the adapters in the early and mid parts based only on the loss from the second exit. The last exit is equivalent to `A. Paral' \cite{rebuffi2018efficient}: we used the loss from the last exit to learn the adapters throughout the base model. In `No-adp', we only used the base network and we trained MLP exit modules with one layer of size 128. `CL' is an experiment with exit module with a $1 \times 1$ convolutional layer. It outputs the same number of channels of its input.

We present our experimental results in Table \ref{tab:results}. We report the classification accuracy (top-1) for the three exits (early, mid, and late) for the different exit topologies. In Table \ref{tab:results},  ‘\%P.’ refers to the total number of parameters required for all the ten domains as a factor of the base model pretrained on ImageNet, we report classification accuracy (top-1) for all the domains of the benchmark from `ImNet' to `UCF', and the last column is the mean accuracy. `Finetune' \cite{rebuffi2018efficient} row reports the results finetuning the base network for every individual domain. `A. Series' reports the published results \cite{rebuffi2017learning} for the series configuration of the adapters within the base network.

The `MLP128' scores a mean accuracy of 62.35\% across all the domains from exit 1 and 71.48\% from exit 2 outperforming all the other variants of the exit modules with MLPs. `MLP128' also performs better than the basic exit module. It achieves a mean gain of $\sim$2\% in the first exit and $\sim$1.17\% in the second exit. The extra dense layers help the network by giving the exit branches some classification enhancement. On the other hand, exit module with a convolutional layer `CL' was not able to learn any extra domain specific features, and it performs as well as the basic exit module.

We see that our proposed training approach results in an accuracy gain of 0.81\% in the first exit and 1.24\% in the second exit compared to block-wise training. This confirms that it is beneficial to train using the losses of the full network also when an early exit is used for inference.

Using the results from the different exits, we can categorize domains into three different levels of difficulty: easy, intermediate, and challenging. Here, we set $T_n=3.5\%$, i.e., for each dataset we selected the earliest exit, where the accuracy loss compared to the performance that we achieve on individual domains if all of full adapted base network was used is no higher than $T_n$. We selected this threshold as a compromise between accuracy and model size. In other applications, different threshold values can be selected based on the accuracy requirements and/or computational/memory resources. The best performance across exit modules within the 3.5\% threshold is shown in the last row 'Best T=3.5\%' in Table \ref{tab:results}. This row illustrates the capability of the proposed method to produce adapters with varying complexity for different domains according to their complexity, so that the overall classification accuracy remains high, while the number of parameters is reduced.

Within the easy category fall DTD and GTSR. We achieved 87.72\% and 97.00\% with accuracy losses of 2.85\% and 2.2\% on DTD and GTSR, respectively, using only one third of the base network with its corresponding domain specific adapters. 
For the intermediate difficulty level, we find  DTD, OGlt, and SVH. We achieved 49.15\%, 87.13\% and 95.35\% in accuracy, respectively. The accuracy loss for DTD was 2.5\%, for OGlt it was 0.29\%, and for SVH it was 0.49\% using only two thirds of the base network and the domain-specific adapter. 
The most challenging domains were: Aircraft 50.29\%, C100  81.01\%, Flwr 70.24\% and UCF 48.64\% which require the parameterization of the whole network.

Allowing a small loss in accuracy, i.e., using a larger threshold $T_n$, we can further reduce the number of parameters. An accuracy of 90.16\% is achieved on SVHN  with a 5.68\% accuracy loss by utilizing only one third of the network. In the same manner, we achieve 43.33\% accuracy on UCF with a 4.68\% accuracy loss with only two thirds of the network.

When the base network is used with no adapters, the performance decreases throughout the network exits (31.71\% in Exit-1, 27.37\%  in Exit-2, and 22.83\%  in Exit-3). All three exits perform poorly on the different domains. The first exit performs better than the later ones since it is easier to use basic features to classify the visually different images. The deeper we go in the network, the more the features become ImageNet-specific and complex. Without adaptation, these features can not be used to extract useful information for domain specific classification tasks. The non-adapted network performed better on domains that are close to ImageNet compared to dissimilar ones (e.g., on DPed, the pedestrian detection task, it has a decent performance as also ImageNet has images with people).

Fig. \ref{gtsrb} presents the learning behavior of an easy domain, in this case GTSRB. We can observe that the three exits converge to the same point around 99\% for accuracy and almost 0 for the loss. In this case, we can achieve the same accuracy level by adapting only the early block of the network. Fig. \ref{svhn} shows the learning behavior on the SVHN dataset considered as intermediate difficulty. This time the second and the third exits are converging toward the same point and the first exit is lagging behind. A comparable performance can be reached using only the early and mid blocks of the network, the early block alone does not have sufficient capacity and learned features to obtain a competitive performance. Finally, Fig. \ref{vgg} shows a case of the challenging domain which requires the whole network to achieve good performance. We can see a clear gap in the performance between the three different exits.

\section{Conclusions}
In this paper, we proposed an adaptive method for incremental multi-domain learning for reducing the required base model parameters and the domains specific parameters added for domains with different levels of complexity. 
We showed through experimentation that the same levels of performance can be achieved with only a portion of the base network and its corresponding adapters for some less complex domains. Our approach encourages an efficient adaptation of multi-domain paradigms using domain agnostic parameters.

\section*{Acknowledgement}
This work was supported by Business Finland and NSF CVDI project AMALIA co-funded by Dead Set Bit and TietoEvry. Special thanks to Pauli Ervi (Dead Set Bit) and Tomi Teikko and Matti Vakkuri (TietoEvry) for their support.

\bibliographystyle{IEEEtran}
\bibliography{IEEEabrv.bib}

\end{document}